\documentclass[11pt,letterpaper]{llncs}
\usepackage{amsmath}
\usepackage{amssymb}
\usepackage{graphicx}
\usepackage{svg}
\usepackage{marvosym}
\usepackage{url}
\usepackage{fancyhdr}
\usepackage{caption}

\usepackage{geometry}
\newcommand{\keywords}[1]{\par\addvspace\baselineskip
\noindent\keywordname\enspace\ignorespaces#1}

\fancypagestyle{firstpage}{\fancyhf{}
}

\begin{document}

\title{\LARGE{ ParaFusion: A Large-Scale LLM-Driven English Paraphrase Dataset Infused with High-Quality Lexical and Syntactic Diversity }}

\author{\large{Lasal Jayawardena \and Prasan Yapa}}
\institute{\large{School of Computing, Informatics Institute of Technology,\\ Colombo 00600, Sri Lanka}}

\maketitle

\thispagestyle{firstpage}

\begin{abstract}
Paraphrase generation is a pivotal task in natural language processing (NLP). Existing datasets in the domain lack syntactic and lexical diversity, resulting in paraphrases that closely resemble the source sentences. Moreover, these datasets often contain hate speech and noise, and may unintentionally include non-English language sentences. This research introduces ParaFusion, a large-scale, high-quality English paraphrase dataset developed using Large Language Models (LLM) to address these challenges. ParaFusion augments existing datasets with high-quality data, significantly enhancing both lexical and syntactic diversity while maintaining close semantic similarity. It also mitigates the presence of hate speech and reduces noise, ensuring a cleaner and more focused English dataset. Results show that ParaFusion offers at least a 25\% improvement in both syntactic and lexical diversity, measured across several metrics for each data source. The paper also aims to set a gold standard for paraphrase evaluation as it contains one of the most comprehensive evaluation strategies to date. The results underscore the potential of ParaFusion as a valuable resource for improving NLP applications.
\end{abstract}

\section{Introduction}

Paraphrase Generation, also known as Question Paraphrase Generation, is a fundamental task and a significant area of focus in NLP. This field has been the subject of research for several decades. Paraphrase Generation plays a critical role in data augmentation, a process that is vital for enhancing the performance of numerous NLP tasks. By generating diverse expressions of identical information, it significantly enriches the training data, thereby improving the robustness and generalization capabilities of NLP models \cite{mckeown_paraphrasing_1979}\cite{meteer_strategies_1988}\cite{kozlowski_generation_2003}\cite{hutchison_synonymous_2004}.

In recent years, neural-based approaches, such as sequence-to-sequence models, have been increasingly used for paraphrase generation due to their ability to learn complex patterns and generate fluent text. However, they require large amounts of high-quality annotated data for training, which can be difficult and costly to obtain. Data quality determines the capability of the model to generate diverse paraphrases. Existing models often struggle with maintaining the semantic equivalence between the original text and the generated paraphrase, especially for longer and more complex sentences. These limitations stem from the issue that the current datasets that are available do not have high-quality paraphrases. A paraphrase to be considered a high-quality paraphrase needs to be lexically diverse, syntactically diverse, grammatically correct, and semantically similar. Section \ref{sec:evaluation-section} outlines an analysis of existing data sources that highlight this issue. Apart from dataset quality, there is no proper evaluation strategy employed by researchers that assesses the quality of diverse paraphrases. Most existing work primarily use lexical metrics to determine model quality whereas the other three components of high-quality paraphrases are often ignored \cite{zhou_paraphrase_2021}.

In light of these challenges, ParaFusion is introduced as a more precise English Dataset to address these issues. LLMs have gained a lot of attraction in recent years, significantly outperforming several state-of-the-art models (SOTA) in several domains \cite{openai_gpt-4_2023}. In this paper, we employ existing datasets and augment the data to generate high-quality paraphrases using the ChatGPT (gpt-3.5-turbo) LLM to create ParaFusion. This paper provides a comprehensive analysis of ParaFusion, investigating it using a range of evaluation metrics to explore various facets of the dataset's quality,  as shown in Section \ref{sec:evaluation-section}, to demonstrate that the paraphrases generated by ParaFusion are more diverse in terms of syntax and lexical compared to existing datasets, while simultaneously maintaining strong semantic similarity between paraphrased sentences. We utilize a rigorous framework for dataset evaluation in hopes of setting a gold standard for future paraphrase evaluation research. Such a comprehensive strategy is needed to precisely evaluate paraphrases which is drastically different from other Sequence-to-Sequence NLP Taks.  Moreover, our human evaluation results Section \ref{sec:human-eval} corroborate that ParaFusion indeed offers higher-quality paraphrases compared to previous datasets. The results suggest a notable potential for realizing enhancements in paraphrase generation tasks, underlining ParaFusion’s ability to shepherd future advancements in NLP.

\section{Related Work}

The landscape of paraphrase generation datasets is critical to understanding the research context and challenges in this domain. This section presents a review of noteworthy paraphrase datasets, highlighting their strengths and limitations.


\textbf{Paraphrase Database (PPDB)} PPDB is a comprehensive resource that houses over 220 million paraphrase pairs \cite{ganitkevitch_ppdb_2013}. The PPDB is compiled through a technique known as bilingual pivoting. The rationale behind this approach is that if two English phrases are translated into the same foreign language phrase, they can be inferred to have identical meanings. Each pair within the PPDB is accompanied by a range of scores, such as paraphrase probabilities and monolingual distributional similarity scores. However, despite its extensive content and detailed scoring system, the PPDB's utility has been questioned recently due to its exclusive focus on phrasal and lexical paraphrases, neglecting sentence paraphrases.

\textbf{Twitter URL}  The Twitter URL dataset \cite{lan_continuously_2017} is a comprehensive collection of large-scale sentential paraphrases sourced from Twitter and connected through shared URLs. This dataset is bifurcated into two subsets, each encompassing both paraphrases and non-paraphrases. The labeling of one subset is performed by human annotators, while the other subset is labeled automatically. It should be noted that the annotation does contain some noise due to the automatic labeling of sentence pairs. Due to the noisiness of the labels, this dataset is not widely used.

\textbf{Wiki Answer} The Wiki Answer dataset \cite{fader_paraphrase-driven_2013} encompasses an estimated 18 million pairs of questions that are paraphrased. The dataset was constructed by mapping open-domain questions to queries over a database of web extractions. The dataset also includes word alignments that connect synonyms within the paraphrased sentences. The dataset is limited in scope as all the sentences provided are in the form of questions, thereby confining the paraphrases to question format only. The dataset is also noisy such that paraphrases do not have the needed semantic similarity of a high-quality dataset.

\textbf{MSCOCO} The MSCOCO dataset \cite{fleet_microsoft_2014} was primarily characterized as a comprehensive object detection dataset. It comprises over 120,000 images, each of which is accompanied by five distinct captions, contributed by five separate annotators.  Typically, the annotators focus on detailing the most conspicuous object or action within an image, rendering this dataset particularly useful for tasks related to paraphrasing.

\textbf{Microsoft Research Paraphrase Corpus} The Microsoft Research Paraphrase Corpus (MRPC) Dataset \cite{dolan_automatically_2005} comprises 5800 sentence pairs derived from online news sources. It also includes human annotations that denote whether each pair represents a paraphrase or semantic equivalence relationship. This was one of the oldest datasets which is still being used for model evaluation but its only downside is that there are very few sentences in the corpora. The sentence length is quite longer compared to other phrase heavy datasets making it a valuable addition.

\textbf{Quora} The Quora Dataset or Quora Question Pair Dataset \cite{iyer_first_2017} which is predominantly used for training and evaluation, contains 150,000 question pairs that are annotated as paraphrases. These validated paraphrase question pairs were specifically employed for the training and testing phases of the paraphrase generation task. The dataset is similar to WikiAnswer in its limitation of having only questions.

\textbf{ParaNMT} The ParaNMT dataset \cite{wieting_paranmt-50m_2018} comprises over 50 million pairs of English sentential paraphrases. These pairs were autonomously generated through the application of back-translation to translate the non-English component of a substantial Czech-English parallel corpus. A Czech-English Neural Machine Translation (NMT) system was employed to translate Czech sentences from the training data into English. These translations were then paired with the English references to form English-English paraphrase pairs. This is the first paraphrase dataset that utilized back-translation. Upon analysis, one downside is the inclusion of improperly formed paraphrases and non-English sentences.

\textbf{ParaBank} The ParaBank datasets \cite{hu_parabank_2019,hu_large-scale_2019} were developed using a Czech-English Neural Machine Translation (NMT) system to generate new paraphrases of English reference sentences. The first version, ParaBank1, introduced lexical constraints to the NMT decoding process, allowing for the generation of multiple high-quality sentential paraphrases for each source sentence. This resulted in an English paraphrase resource that exhibits a higher degree of lexical diversity. Its successor, ParaBank2, addressed the issue of syntactic diversity by providing multiple diverse sentential paraphrases. These paraphrases were generated from a bilingual corpus using negative constraints, inference sampling, and clustering. Even with the improvements both datasets still lack syntactic diversity.

\textbf{PAWS (Paraphrase Adversaries from Word Scrambling)} The PAWS Dataset \cite{zhang_paws_2019} contains sentences with high bag-of-words (BOW) overlap but having different word order. The PAWS dataset creation involved a two-step process. Initially, a language model was used to generate sentence pairs with high lexical overlap through word swapping, ensuring naturalness and well-formedness. Subsequently, back translation was employed to create paraphrases with high bag-of-words overlap but distinct word order. The PAWS dataset is further divided into two subsets: PAWSQQP and PAWSWiki. The PAWSQQP subset is derived from the Quora Question Pairs (QQP) corpus, while the PAWSWiki subset is derived from Wikipedia. Subsets of the dataset were then subjected to human review for sentence correction and paraphrase identification. PAWSWiki has significantly better quality paraphrases than all the other datasets, yet improvements can be made for syntactic diversity. There is a considerable portion of PAWS that is noisy and the other portion that is labeled exhibit high-quality.

\section{ParaFusion}

\begin{figure}
\centering
\includegraphics[width=0.95\textwidth]{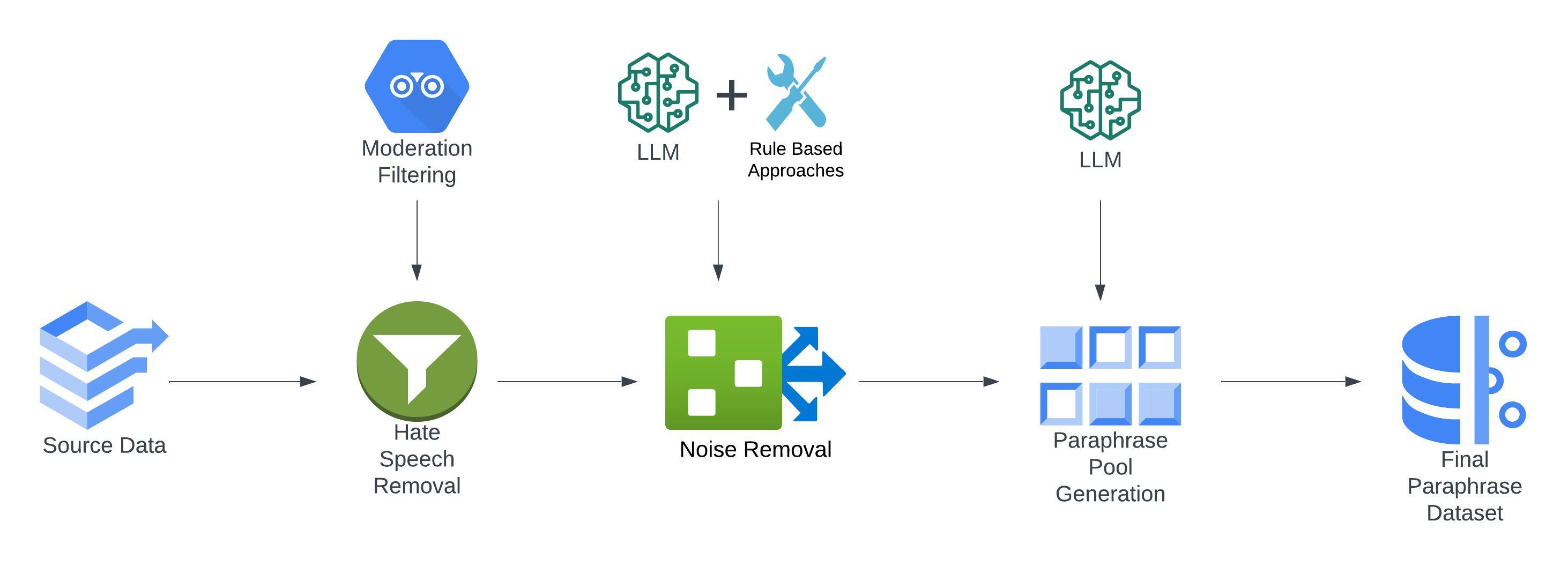}
\caption{High-level diagram outlining the dataset creation process.}
\label{fig:high-level-diagram}
\end{figure}

\subsection{Data Sources}

ParaFusion is a comprehensive tool constructed on the foundation of several datasets, each contributing unique elements to the overall structure. The first dataset we utilized was the MRPC Dataset \cite{dolan_automatically_2005}, which provided a solid base for our work. Following this, we incorporated a subset of the Quora Dataset \cite{iyer_first_2017}, specifically selecting sentences that were labeled as paraphrases to enrich our data pool. To further diversify our data, we included PAWSWiki, a component of the PAWS Dataset \cite{zhang_paws_2019}. However, we consciously decided against using PAWSQQP as it contained source sentences identical to those in the Quora Dataset, which would have introduced unnecessary redundancy into our data. We also considered the use of three additional datasets: ParaNMT \cite{wieting_paranmt-50m_2018}, Parabank1 \cite{hu_parabank_2019}, and Parabank2 \cite{hu_large-scale_2019}. However, due to financial constraints, we were unable to utilize these datasets in their entirety. Instead, we strategically selected all the common source sentences from these datasets and supplemented this with an additional 250,000 source sentences. 

The decision to incorporate multiple datasets into our research methodology was a strategic one, aimed at enhancing the comprehensiveness and diversity of the dataset. Different datasets inherently possess varying sentence lengths and unique content, style, and context. By amalgamating several datasets, we ensured that ParaFusion captured a broad spectrum of sentence lengths, and provided an extensive range of topics, writing styles, and contexts for training. This approach also helped us to mitigate the risk of potential data bias, ensuring a more balanced and representative dataset.

It should be emphasized that the method we suggest is not confined to these particular datasets. Instead, it has the potential to be utilized with any general text to generate paraphrases.

\subsection{Base Dataset Creation}

In the construction of ParaFusion, we initially amalgamated the aforementioned data sources, selecting approximately 750,000 source sentences for paraphrase generation. An initial pass was conducted to filter out offensive content, utilizing OpenAI’s Moderation Endpoint \cite{openai_moderation_2023}. This process flagged any source sentence that fell under the categories of "sexual", "hate", "harassment", "self-harm", "sexual/minors", “hate/threatening", "violence/graphic", "self-harm/instructions", "self-harm/intent", "harassment/threatening", or "violence". This enabled us to filter out approximately 50,000 source sentences containing offensive content. 

Subsequently, we employed the ChatGPT (gpt-3.5-turbo) LLM to augment the source sentences. The prompt used for each dataset varied slightly, and the reasons for these variations are discussed in the Additional Processing Section \ref{sec:additonal-processing}. See Figure \ref{fig:prompt-example} for an example illustration of this.

\begin{figure}
\centering
\noindent\fbox{%
    \begin{minipage}{0.99\textwidth}
    Given a Source Sentence: ```\$Source Sentence```, generate 5 diverse paraphrases.  Try to generate paraphrases that are both lexical and syntactically diverse from the Source Sentence. Give the output as a numbered list.
    \end{minipage}%
}
\caption{This figure illustrates a sample prompt fed to the gpt-3.5-turbo model for generating diverse paraphrases.}
\label{fig:prompt-example}
\end{figure}

We adopted an iterative prompt engineering approach to construct an effective prompt. This prompt was then input into the gpt-3.5-turbo model, along with the source sentences. For paraphrase generation, we set the temperature parameter to 0. The model output was a string containing augmented, diverse sentence paraphrases in a numbered list format. This string was subsequently processed to generate a list of strings, which was then used to construct the dataset. By utilizing the source sentences, we successfully generated nearly 3.5 million sentence paraphrases.

\subsection{Additional Processing}
\label{sec:additonal-processing}

Our study revealed that datasets constructed using back-translation \cite{wieting_paranmt-50m_2018}\cite{hu_parabank_2019}\cite{hu_large-scale_2019} often contained a significant amount of noise, including non-English source sentences. To tackle this, we iteratively developed a new prompt for the gpt-3.5-turbo model, instructing it to identify English sentences and generate paraphrases, or output "Error" for non-English sentences. This method, coupled with a rule-based approach to filter out certain responses from the model, proved efficient in significantly reducing noise, and eliminating approximately 10,000 non-English source sentences during generation.

In the final stage of dataset creation, we didn't merely use the source sentence and generated paraphrase pairs as they were. Instead, we treated the source sentence and generated paraphrase as a pool of sentences, from which we created unique paraphrase pairs. This approach diversified the paraphrase pairs in our dataset, ensuring a wider range of sentence structures and expressions. 

This method offers several benefits. Firstly, it enhances the diversity of our dataset, which is crucial for training robust and generalizable models. Secondly, it helps to mitigate the risk of overfitting by exposing our models to a more representative sample of the data they will encounter in the real world. Lastly, it helps to reduce the impact of any noise in the source sentence. By creating unique paraphrase pairs, we can ensure that any noise in the source sentences or generated paraphrases are not consistently paired with the same sentences, thereby reducing the likelihood that our models will learn to associate this noise with specific inputs or outputs. 

The final dataset comprises around 2 million unique paraphrase sentence pairs. These techniques significantly reduced noise in ParaFusion, contributing to the construction of a higher-quality dataset, comparable to human annotation or the next best alternative.

\section{Evaluation}
\label{sec:evaluation-section}

In this section, we evaluate ParaFusion, which was created using several source datasets: the Microsoft Research Paraphrase Corpus (MRPC), a subset of the Quora Dataset, PAWSWiki, ParaNMT, Parabank1, and Parabank2. For a fair evaluation, we only consider source sentences common to both ParaFusion and these source datasets. 

We use "Para-Common Subset" to denote paraphrases in ParaFusion common to ParaNMT, Parabank1, and Parabank2. "MRPC Subset", "PAWS Subset", and "QQP Subset" refer to paraphrases in ParaFusion common to MRPC, Quora Dataset, and PAWSWiki Dataset, respectively. Each subset is analyzed separately for fairness.

\subsection{Quantitative Analysis}

We adopt a comprehensive quantitative evaluation methodology to assess the data sources and sentence pairs in ParaFusion. To guarantee a fair evaluation, the dataset is partitioned into four segments based on the source sentences. Our evaluation focuses on three crucial characteristics: semantic similarity, syntactic diversity, and lexical diversity.

\subsubsection{Semantic Similarity}

Semantic similarity is a measure of the degree to which two pieces of text are related in terms of their meaning. In our research, we quantify this similarity by leveraging various models to obtain sentence embeddings of the source and the paraphrase, and then calculating the cosine similarity between these embeddings. 

For instance, we use the "Ada Score" which is derived from OpenAI’s text-embedding-ada-002 model \cite{openai_new_2023}. Similarly, the "SimCSE Score" is calculated using SimCSE’s sup-simcse-roberta-large model \cite{gao_simcse_2021}, and the "PromCSE Score" is based on PromCSE’s sup-promcse-roberta-large model \cite{jiang_improved_2022}. 

We also utilize several models from the sentence-transformers library \cite{reimers_sentence-bert_2019}. The "Mpnet Score" is calculated using the all-mpnet-base-v2 model, while the "Mpnet-qa Similarity Score" is derived from the multi-qa-mpnet-base-dot-v1 model. The "Roberta Score" is based on the all-distilroberta-v1 model, and the "Mini Score" and "Mini Score2" are calculated using the all-MiniLM-L12-v2 and all-MiniLM-L6-v2 models, respectively.

The comprehensive evaluation of semantic similarity is presented in Table \ref{tab:semantic-comparison}. It is evident that ParaFusion not only maintains semantic similarity but also, in some cases, surpasses the quality of the original data source. The perceived low similarity in less sophisticated models can be attributed to the lack of lexical or syntactic diversity in the original source sentences, which results in highly similar sentences. However, in ParaFusion, the complexity of the sentences allows for a more accurate measurement of semantic similarity using advanced models such as the text-embedding-ada-002 or SIMCSE. 

\begin{table*}
\hspace*{-1cm}
\centering
\begin{tabular}{|p{3.3cm}|p{1.4cm}|p{1.4cm}|p{1.7cm}|p{1.4cm}|p{1.6cm}|p{1.4cm}|p{1.4cm}|p{1.4cm}|}
\hline
\textbf{Data Source} & \textbf{Ada Score (↑)} & \textbf{SimCSE Score (↑)} & \textbf{PromCSE Score (↑)} & \textbf{Mpnet Score (↑)} & \textbf{Mpnet-qa Score (↑)} & \textbf{Roberta Score (↑)} & \textbf{Mini Score (↑)} & \textbf{Mini Score2 (↑)} \\
\hline
MSR Original & 95.53\% & 86.50\% & 99.22\% & 84.27\% & 86.23\% & 82.80\% & 83.27\% & 82.47\% \\
MSR Subset (Ours) & \textbf{96.59\%} & \textbf{93.30\%} & \textbf{99.56\%} & \textbf{88.33\%} & \textbf{90.80\%} & \textbf{86.45\%} & \textbf{86.82\%} & \textbf{86.01\%} \\
\hline
QQP Original & \textbf{95.52\%} & 89.61\% & \textbf{99.65\%} & \textbf{88.74\%} & \textbf{91.30\%} & \textbf{87.00\%} & \textbf{88.75\%} & \textbf{88.37\%} \\
QQP Subset (Ours) & 94.56\% & \textbf{89.87\%} & \textbf{99.65\%} & 86.86\% & 89.33\% & 84.29\% & 85.12\% & 83.85\% \\
\hline
PAWS Original & \textbf{98.90\%} & \textbf{97.42\%} & \textbf{99.90\%} & \textbf{96.96\%} & \textbf{97.06\%} & \textbf{97.20\%} & \textbf{97.24\%} & \textbf{97.13\%} \\

PAWS Subset (Ours) & 96.22\% & 92.20\% & 99.34\% & 91.74\% & 91.57\% & 91.72\% & 92.08\% & 91.33\% \\
\hline
ParaNMT Original & 94.83\% & 87.50\% & 99.88\% & 87.00\% & 90.67\% & 82.05\% & 87.69\% & 87.58\% \\
ParaBank1 Original & \textbf{95.58\%} & \textbf{89.85\%} & \textbf{99.91\%} & \textbf{89.00\%} & \textbf{92.09\%} & \textbf{85.31\%} & \textbf{89.60\%} & \textbf{89.51\%} \\
ParaBank2 Original & 94.49\% & 85.52\% & 98.87\% & 82.89\% & 87.60\% & 79.76\% & 83.63\% & 83.26\% \\
Para-Common Subset (Ours) & 95.04\% & 74.32\% & 98.60\% & 67.16\% & 74.11\% & 63.00\% & 67.19\% & 65.80\% \\
\hline
\end{tabular}

\caption{Comparison of Semantic Similarity of the original data sources and corresponding ParaFusion data subsets.}
\label{tab:semantic-comparison}
\end{table*}

\subsubsection{Syntactic Diversity}

Syntactic diversity refers to the variety and complexity of sentence structures of a paraphrase given a source sentence. High syntactic diversity indicates that the paraphrase sentences are diverse and linguistically rich. We assess this diversity using several metrics. 

The "Ted-F Score" and "Ted-3 Score" are calculated by building constituency parse trees for the source and paraphrase sentences using Stanza \cite{qi_stanza_2020}, converting the trees to bracket notation using the NLTK library \cite{bird_natural_2009} and regex, and then using the APTED library \cite{pawlik_efficient_2015} to calculate the Full Tree Edit Distance and the Tree Edit Distance of the first three layers, respectively. 

The "Kermit Score" is calculated by obtaining the cosine similarity of the source and the paraphrase syntactic embeddings using the Kermit library \cite{zanzotto_kermit_2020}, and then subtracting this similarity from one. 

The "ST Kernel Score" and "NP Kernel Score" are calculated by first building constituency parse trees for the source and paraphrase sentences using Stanza, converting the trees to an NLTK Tree, and then finding all the subtrees or node pairs, respectively. Then the kernel similarity is calculated using the number of unique common subtrees or node pairs over the number of unique total subtrees or node pairs, and then subtracting this similarity from one. "ST Kernel Score" is the score calculated related to the subtrees in the constituency parse trees whereas "NP Kernel Score" is the score calculated using the node pair similarity.

The full syntactic diversity evaluation is shown in Table \ref{tab:syntactic-comparison}. We can see that ParaFusion has a significant improvement over all the original data sources. This is because it was able to generate more syntactically rich paraphrases. 

\begin{table*}
\centering
\begin{tabular}
{|p{4.5cm}|p{1.8cm}|p{1.8cm}|p{2cm}|p{2.cm}|p{2cm}|}
\hline
\textbf{Data Source} & \textbf{Ted-F Score (↑)} & \textbf{Ted-3 Score (↑)} & \textbf{Kermit Score (↑)} & \textbf{ST Kernel Score (↑)} & \textbf{NP Kernel Score (↑)} \\
\hline
MSR Original & 18.65 & 3.76 & 55.23\% & 65.18\% & 82.87\% \\
MSR Subset (Ours) & \textbf{29.09} & \textbf{4.92} & \textbf{64.37\%} & \textbf{73.55\%} & \textbf{89.34\%} \\
\hline
QQP Original & 9.30 & 2.02 & 57.59\% & 70.14\% & 88.29\% \\
QQP Subset (Ours) & \textbf{16.91} & \textbf{3.35} & \textbf{71.62\%} & \textbf{81.13\%} & \textbf{93.64\%} \\
\hline
PAWS Original & 9.46 & 1.70 & 37.74\% & 47.44\% & 73.47\% \\
PAWS Subset (Ours) & \textbf{27.02} & \textbf{4.67} & \textbf{61.06\%} & \textbf{69.48\%} & \textbf{87.35\%} \\
\hline
ParaNMT Original & 3.12 & 2.01 & 54.06\% & 73.99\% & 82.51\% \\
ParaBank1 Original & 2.06 & 1.48 & 45.85\% & 63.22\% & 70.37\% \\
ParaBank2 Original & 3.76 & 2.07 & 61.94\% & 85.01\% & 96.78\% \\
Para-Common Subset (Ours) & \textbf{9.33} & \textbf{3.99} & \textbf{80.78\%} & \textbf{91.73\%} & \textbf{97.94\%} \\
\hline
\end{tabular}

\caption{Comparison of Syntactic Diversity of the original data sources and corresponding ParaFusion data subsets.}
\label{tab:syntactic-comparison}
\end{table*}

\subsubsection{Lexical Diversity}

Lexical diversity refers to the range and variety of words used in a text. It is a measure of the breadth of vocabulary and the use of synonyms. In the context of paraphrasing, assessing lexical diversity is crucial to understand the extent of vocabulary variation. We used several metrics to assess lexical diversity.

The "BOW Overlap" is calculated by determining the intersection of tokens between the source and the paraphrase, divided by the total number of tokens. This value is then subtracted from one. 

The "Corpus BLEU" and "Corpus BLEU2" scores are calculated using the SacreBLEU Library \cite{post_call_2018}. The Corpus BLEU2 score uses the “method1” smoothing function in the SacreBLEU library. Both scores are then subtracted from one.

The "Sentence BLEU" score is calculated in a similar manner to the Corpus BLEU score using the SacreBLEU Library but at the sentence level. This score is also subtracted from one.

The "METEOR" score is calculated using the NLTK library and then subtracted from one.

The "ROUGE 1", "ROUGE 2", and "ROUGE L" scores are calculated using the Google Research library \cite{google_google_2023} and then subtracted from one.

The "Token $\cap / \cup$" score is similar to the BOW Overlap score, but with a small difference. It is calculated using the intersection of tokens between the source and the paraphrase, divided by the total number of unique tokens. This value is then subtracted from one.

The "Google BLEU" score is calculated using Huggingface’s Evaluate library and then subtracted from one.\footnote{{\label{footnote1}https://github.com/huggingface/evaluate}}

The "TER (Translation Error Rate)", "WER (Word Error Rate)", and "CharacTER (Character Error Rate)" scores are calculated using Huggingface’s Evaluate library.\textsuperscript{\ref{footnote1}}

The full lexical diversity evaluation is shown in Table \ref{tab:lexical-comparison-1} and Table \ref{tab:lexical-comparison-2}. We can see that ParaFusion has a significant improvement over all the original data sources. 

\begin{table*}
\hspace*{-1cm}
\centering
\begin{tabular}{|p{3.3cm}|p{1.9cm}|p{1.7cm}|p{1.8cm}|p{2cm}|p{2.4cm}|p{2cm}|}
\hline
\textbf{Data Source} & \textbf{1 - BOW Overlap (↑)} & \textbf{1 - Corpus BLEU (↑)} & \textbf{1 - Corpus BLEU2 (↑)} & \textbf{1 - Sentence BLEU (↑)} & \textbf{1 - METEOR (↑)} & \textbf{1 - ROUGE 1 (↑)} \\
\hline
MSR Original & 35.37\% & 93.92\% & 99.62\% & 59.75\% & 31.43\% & 29.49\% \\
MSR Subset (Ours) & \textbf{43.72\%} & \textbf{96.53\%} & \textbf{99.65\%} & \textbf{75.93\%} & \textbf{38.96\%} & \textbf{36.81\%} \\
\hline
QQP Original & 36.69\% & \textbf{95.44\%} & 99.15\% & 70.80\% & 35.33\% & 33.46\% \\
QQP Subset (Ours) & \textbf{53.04\%} & 82.21\% & \textbf{99.46\%} & \textbf{84.33\%} & \textbf{50.20\%} & \textbf{51.86\%} \\
\hline
PAWS Original & 19.33\% & 94.21\% & 99.60\% & 34.96\% & 8.58\% & 5.96\% \\
PAWS Subset (Ours) & \textbf{40.26\%} & \textbf{98.15\%} & \textbf{99.65\%} & \textbf{72.39\%} & \textbf{34.59\%} & \textbf{30.22\%} \\
\hline
ParaNMT Original & 53.56\% & 62.19\% & 97.75\% & 58.26\% & 27.33\% & 18.37\% \\
ParaBank1 Original & 46.50\% & 62.47\% & 97.73\% & 51.62\% & 24.54\% & 16.08\% \\
ParaBank2 Original & 55.36\% & 63.06\% & 97.86\% & 66.85\% & 39.24\% & 30.43\% \\
Para-Common Subset (Ours)  & \textbf{74.31\%} & \textbf{82.24\%} & \textbf{98.72\%} & \textbf{86.45\%} & \textbf{69.03\%} & \textbf{64.82\%} \\
\hline
\end{tabular}
\caption{Comparison of Lexical Diversity of the original data sources and corresponding ParaFusion Data Subsets.}
\label{tab:lexical-comparison-1}
\end{table*}

\begin{table*}
\hspace*{-1cm}
\centering
\begin{tabular}{|p{3.3cm}|p{2cm}|p{2cm}|p{1.5cm}|p{1.3cm}|p{1.3cm}|p{1.3cm}|p{1.8cm}|}
\hline
\textbf{Data Source} & \textbf{1 - ROUGE 2 (↑)} & \textbf{1 - ROUGE L (↑)} & \textbf{1 - Token $\cap / \cup$ (↑)} & \textbf{TER (↑)} & \textbf{WER (↑)} & \textbf{CER (↑)} & \textbf{1 - Google BLEU (↑)} \\
\hline
MSR Original & 47.55\% & 33.97\% & 44.95\% & 70.34 & 76.07 & 42.98\% & 55.96\% \\
MSR Subset (Ours) & \textbf{63.22\%} & \textbf{51.86\%} & \textbf{54.79\%} & \textbf{85.35} & \textbf{98.47} & \textbf{64.95\%} & \textbf{69.81\%} \\
\hline
QQP Original & 57.72\% & 36.85\% & 49.04\% & 56.56 & 62.77 & 44.74\% & 64.96\% \\
QQP Subset (Ours) & \textbf{76.95\%} & \textbf{57.89\%} & \textbf{66.33\%} & \textbf{69.92} & \textbf{75.34} & \textbf{76.94\%} & \textbf{79.12\%} \\
\hline
PAWS Original & 21.44\% & 12.78\% & 15.74\% & 14.81 & 23.63 & 14.19\% & 31.80\% \\
PAWS Subset (Ours) & \textbf{56.97\%} & \textbf{47.10\%} & \textbf{66.33\%} & \textbf{56.10} & \textbf{85.66} & \textbf{62.38\%} & \textbf{65.92\%} \\
\hline
ParaNMT Original  & 33.20\% & 19.25\% & 61.39\% & 30.10 & 66.10 & 25.74\% & 44.99\% \\
ParaBank1 Original & 31.69\% & 16.90\% & 52.51\% & 23.71 & 50.47 & 21.61\% & 37.37\% \\
ParaBank2 Original & 55.24\% & 32.97\% & 67.69\% & 46.84 & 68.01 & 35.39\% & 66.87\% \\
Para-Common Subset (Ours) & \textbf{89.12\%} & \textbf{69.68\%} & \textbf{82.23\%} & \textbf{82.46} & \textbf{89.31} & \textbf{85.10\%} & \textbf{81.30\%} \\
\hline
\end{tabular}
\caption{Comparison of Lexical Diversity of the original data sources and corresponding ParaFusion Data Subsets.}
\label{tab:lexical-comparison-2}
\end{table*}

\subsection{Qualitative Evaluation}

During the qualitative analysis, we uncovered intriguing information. A prevalent issue in existing datasets is that most paraphrases are merely sentences with substituted synonyms. Figure \ref{fig:qualitative-1} illustrates an example of this, where "A" is replaced by "One", leaving the rest of the sentence identical to the source. This does not constitute an effective paraphrase. 

Another problem we observed is that existing data sources, particularly those relying on back-translation, often contain paraphrases with altered meanings due to word choice. Figure \ref{fig:qualitative-2} exemplifies a situation where the meaning of a paraphrase has deviated from the original source datasets. This occurs when inappropriate words are used without considering the context. For instance, replacing "culverts" with "driers" is not suitable in this context. 

In contrast, paraphrases in ParaFusion demonstrate superior lexical diversity and more syntactic changes, while preserving the original meaning.

\begin{figure}[h]
\centering
\noindent\fbox{%
    \begin{minipage}{0.47\textwidth}
    \textbf{Source Sentence}: A poetic example of early modern philosophical thought can be found in the surprising works of the renowned intellectual Stoyan Mihaylovski.
    \newline 
    \textbf{Original Paraphrase}: One poetic example of early modern philosophical thought can be found in the surprising works of the renowned intellectual Stoyan Mihaylovski.
    \newline 
     \textbf{ParaFusion Paraphrase}: Stoyan Mihaylovski's works are a remarkable representation of early modern philosophical thought, expressed in a poetic manner.

    \end{minipage}%
}
\caption{This figure illustrates an instance where the paraphrase 
 in a source dataset has only word substitutions.}
\label{fig:qualitative-1}
\end{figure}

\begin{figure}[h]
\centering
\noindent\fbox{%
    \begin{minipage}{0.47\textwidth}
    \textbf{Source Sentence}: The water is moved by the gravity and is controlled by huge valves in the driers .
    \newline 
    \textbf{Original Paraphrase}: The water is moved by gravity and is controlled by huge valves in the culverts .
    \newline
     \textbf{ParaFusion Paraphrase}: Huge valves in the driers regulate the movement of water, which is facilitated by the force of gravity.

    \end{minipage}%
}
\caption{This figure illustrates an instance where the paraphrase 
 in a previous dataset has a different meaning.}
\label{fig:qualitative-2}
\end{figure}

Furthermore, we observed that the original data sources contained a substantial number of paraphrase pairs where the source and the paraphrase were identical. This presents a significant problem because training a model on such datasets could encourage the model to simply reproduce the input, thereby defeating the purpose of paraphrasing. This qualitative analysis further underscores the effectiveness and reliability of ParaFusion in these scenarios.

\subsection{Human Evaluation}
\label{sec:human-eval}

In our research, we conducted human evaluations using four annotators who assessed approximately 7000 paraphrase pairs across various datasets, including the source and ParaFusion. The source sentences were selected as follows: 200 from the MRPC Dataset, 250 from the Quora Dataset, 250 from the PAWSWiki, and 300 common source sentences from ParaNMT, Parabank1, and Parabank2. The sentences were sampled to ensure that the lengths of the data source sentences were properly represented. The corresponding paraphrases from the source datasets and ParaFusion were then selected, resulting in a total of 7000 paraphrase pairs for evaluation.

For the evaluation, we used a 5-point Likert scale \cite{van_der_lee_best_2019} to assess key metrics, including Semantic Similarity, Lexical Diversity, Syntactic Diversity, and Grammatical Correctness. The full breakdown of the Likert scale can be seen in Figure \ref{fig:human-annotator-instructions}. In this scale, 5 represents the highest level of similarity, diversity, or correctness, while 1 indicates the lowest.

Specifically, semantic similarity ratings ranged from 5 for identical meaning to the source text, to 1 for completely different or unrelated meaning. Lexical diversity was evaluated based on vocabulary range and richness, with 5 indicating excellent diversity and 1 signifying limited diversity. Syntactic diversity was assessed by structural variations, with 5 denoting high diversity and 1 signifying minimal variation. Lastly, grammatical correctness was evaluated, with 5 indicating flawless grammar and 1 representing significant errors impacting comprehension.

Table \ref{tab:human-eval-comparison-small} provides a summary of the human evaluation, and a full breakdown can be seen in Table \ref{tab:human-evaluation-results}. The results clearly indicate that ParaFusion is more lexically and syntactically diverse than the original data sources.

\begin{center}
\begin{tabular}{|p{6.3cm}|p{3.4cm}|p{3.8cm}|}
\hline
 & \textbf{Original} & \textbf{ParaFusion} \\
\hline
\textbf{Semantic Similarity} & 4.36 & 4.46 \\
\textbf{Lexical Diversity} & 2.29 & 3.09 \\
\textbf{Syntactic Diversity} & 2.44 & 3.40 \\
\textbf{Grammatical \newline Correctness} & 4.37 & 4.79 \\
\hline
\end{tabular}
\captionof{table}{Comparison of Semantic Similarity, Lexical Diversity, Syntactic Diversity, and Grammatical Correctness between two data sets in Human Evaluation.}
\label{tab:human-eval-comparison-small}
\end{center}

\subsection{LLM Evaluation}

In the past year, the use of LLMs for evaluation in NLP has gained traction. The primary reason for this trend is the ability of LLMs to outperform existing reference-free metrics \cite{liu_g-eval_2023}. In light of this, we conducted an LLM evaluation using OpenAI's gpt-4 model on the same data provided to our human annotators. The gpt-4 model was selected due to its status as the SOTA LLM at the time of writing this paper \cite{openai_gpt-4_2023}. We designed the prompt using the same instructions given to the human annotators, as illustrated in Figure \ref{fig:llm-eval-prompt}.

The summary of the results is presented in Table \ref{tab:llm-eval-comparison-small}, while a full breakdown is provided in Table \ref{tab:full-gpt4-eval}. The observations clearly indicate that ParaFusion is more lexically and syntactically diverse than the original data sources, mirroring the results of the Human Evaluation.

\begin{center}
\begin{tabular}{|p{6.3cm}|p{3.4cm}|p{3.8cm}|}
\hline
 & \textbf{Original} & \textbf{ParaFusion} \\
\hline
\textbf{Semantic Similarity} & 4.49 & 4.94 \\
\textbf{Lexical Diversity} & 1.75 & 3.34 \\
\textbf{Syntactic Diversity} & 2.02 & 3.84 \\
\textbf{Grammatical \newline Correctness} & 4.75 & 4.99 \\
\hline
\end{tabular}
\captionof{table}{Comparison of Semantic Similarity, Lexical Diversity, Syntactic Diversity, and Grammatical Correctness between two data sets in LLM Evaluation.}
\label{tab:llm-eval-comparison-small}
\end{center}

\section{Conclusion}
This research paper introduces ParaFusion, a large-scale, high-quality English paraphrase dataset developed using LLMs. The dataset is designed to address the limitations of the lack of syntactic and lexical diversity in existing datasets. Additionally, it shows potential as a very high-quality alternative to human-annotated paraphrase pairs which are costly to obtain. ParaFusion augmented existing datasets to generate high-quality data, significantly enhancing both lexical and syntactic diversity while maintaining semantic similarity which was seen in the evaluation section. It also mitigates the presence of hate speech and reduces noise, ensuring a cleaner, more focused English dataset. The evaluation of ParaFusion demonstrated its potential as a valuable resource for improving NLP applications. 

\section*{Limitations}
While ParaFusion addresses several challenges in paraphrase generation, there are a few limitations to be considered. Firstly, the dataset is focused on English paraphrases, which may limit its applicability to other languages. Future research could explore the development of similar datasets for other languages to enhance the diversity and inclusivity of NLP applications. This could be accomplished using multi-lingual LLMs or language-specific LLMs.

Secondly, although efforts were made to ensure the quality and reliability of the dataset, there may still be instances of noise, inaccuracies, or in rare cases, offensive language. The use of LLMs for paraphrase generation introduces the possibility of generating incorrect paraphrases. Researchers and practitioners should exercise caution and conduct thorough evaluations when using the ParaFusion dataset. 

Another point to be considered is the issue of error propagation. Our dataset was generated using gpt-3.5-turbo, thus inheriting all potential risks associated with it. This can also include phenomena like quality drift where the model output can change as the model adapts, if expansions were to be done. This could potentially introduce additional inaccuracies into the paraphrase generation process, and users should be aware of this when using the ParaFusion dataset and expanding it.

Lastly, the evaluation metrics used in this research provide valuable insights into the quality and diversity of the dataset. However, they may not capture all aspects of paraphrase generation. Future research could explore additional evaluation metrics or approaches to further assess the effectiveness and performance of paraphrase generation models using our ParaFusion dataset.

\section*{Ethics Statement}
In conducting this research and developing the ParaFusion dataset, we took several ethical considerations into account. Firstly, we ensured that the dataset contained minimal hate speech or offensive language by implementing a moderation filtering process. This was done to ensure that the dataset is safe and suitable for using in NLP applications.

Secondly, we made efforts to reduce noise in the dataset, such as removing non-English sentences and filtering out responses that did not meet the criteria for high-quality paraphrases. This was done to ensure that the dataset is of high quality and reliable for training NLP models.

Additionally, we considered the potential impact of our research on the broader NLP community. By providing a large-scale, high-quality paraphrase dataset, we aim to contribute to the advancement of NLP applications. This dataset can be used to improve the performance and robustness of NLP models, leading to more accurate and reliable NLP.

\bibliographystyle{IEEEtran}
\bibliography{references}

\vspace{2cm}

\section*{Authors}

\noindent {\bf Lasal Jayawardena} is a fourth-year undergraduate student at the Informatic Institute of Technology (IIT) Sri Lanka, affiliated with Robert Gordon University in Aberdeen, Scotland. He is currently pursuing a Bachelor of Science honors degree in Artificial Intelligence and Data Science. Lasal's research interests primarily revolve around the domain of Natural Language Processing (NLP), with a strong focus on Large Language Models and Deep Learning techniques. \\

\noindent {\bf Prasan Yapa} received Master of Computer Science (Research Major) and Bachelor of Science in Information Technology and Management from University of Moratuwa, Sri Lanka. Currently, he is pursuing his PhD in Computer Science \& Engineering from Kyoto University of Advanced Science in Japan. His research interests include Natural Language Processing, Deep Learning, Affective Computing, Health Informatics and Computer Vision. \\

\newpage

\appendix

\section{Human and LLM Evaluation}
\label{sec:appendix}

\begin{table}
\hspace*{-1cm}
\centering
\begin{tabular}{|p{3.3cm}|p{3.3cm}|p{3cm}|p{3.3cm}|p{3.8cm}|}
\hline
\textbf{Data Source} & \textbf{Semantic Similarity: Scale from 1 to 5} & \textbf{Lexical Diversity: Scale from 1 to 5} & \textbf{Syntactic Diversity: Scale from 1 to 5} & \textbf{Grammatical Correctness: Scale from 1 to 5}\\
\hline
MSR Original & 4.61 & 2.61 & 2.98 & 4.99 \\
MSR Subset (Ours) & \textbf{4.99} & \textbf{3.56} & \textbf{3.91} & \textbf{4.99} \\
\hline
QQP Original & 4.35 & 2.14 & 2.63 & 4.96 \\
QQP Subset (Ours) & \textbf{4.97} & \textbf{3.52} & \textbf{3.82} & \textbf{4.99} \\
\hline
PAWS Original & 4.87 & 1.30 & 1.82 & 4.86 \\
PAWS Subset (Ours) & \textbf{4.95} & \textbf{3.20} & \textbf{3.88} & \textbf{4.98} \\
\hline
ParaNMT Original & 4.49 & 1.34 & 1.41 & 4.78 \\
ParaBank1 Original & 4.66 & 1.34 & 1.40 & 4.81 \\
ParaBank2 Original & 4.07 & 2.07 & 2.26 & 4.27 \\
Para-Common Subset (Ours) & \textbf{4.84} & \textbf{3.08} & \textbf{3.78} & \textbf{5.00} \\
\hline
\end{tabular}
\caption{Full Breakdown of the LLM Evaluation using gpt-4}
\label{tab:full-gpt4-eval}
\end{table}

\begin{table}
\hspace*{-1cm}
\centering
\begin{tabular}{|p{3.3cm}|p{3.3cm}|p{3cm}|p{3.3cm}|p{3.8cm}|}
\hline
\textbf{Data Source} & \textbf{Semantic Similarity: Scale from 1 to 5} & \textbf{Lexical Diversity: Scale from 1 to 5} & \textbf{Syntactic Diversity: Scale from 1 to 5} & \textbf{Grammatical Correctness: Scale from 1 to 5}\\
\hline
MSR Original & 4.26 & 2.66 & 2.96 & 4.80 \\
MSR Subset (Ours) & \textbf{4.53} & \textbf{3.10} & \textbf{3.46} & \textbf{4.88} \\
\hline
QQP Original & 4.25 & 2.38 & 2.62 & 4.92 \\
QQP Subset (Ours) & \textbf{4.49} & \textbf{3.15} & \textbf{3.38} & \textbf{4.96} \\
\hline
PAWS Original & \textbf{4.79} & 2.07 & 2.25 & 4.13 \\
PAWS Subset (Ours) & 4.42 & \textbf{3.02} & \textbf{3.39} & \textbf{4.69} \\
\hline
ParaNMT Original & 4.51 & 2.00 & 2.10 & 4.36 \\
ParaBank1 Original & \textbf{4.66} & 1.97 & 2.04 & 4.24 \\
ParaBank2 Original) & 4.11 & 2.37 & 2.59 & 4.04 \\
Para-Common Subset (Ours) & 4.42 & \textbf{3.02} & \textbf{3.45} & \textbf{4.76} \\
\hline
\end{tabular}
\caption{Full Breakdown of the Human Evaluation}
\label{tab:human-evaluation-results}
\end{table}

\begin{figure}
\centering
\noindent\fbox{%
    \begin{minipage}{0.90\textwidth}
    \textbf{Source Text:} \$source\_text

    \textbf{Paraphrase:} \$paraphrase

    Please evaluate the following aspects of the paraphrase in comparison to its source text on a likert scale of 1 to 5, where:

    \textbf{Semantic Similarity:} This refers to how closely the meaning of the paraphrase matches the meaning of the source text.

    \textit{Rating Scale for Semantic Similairty}

    1: The paraphrase has a completely different meaning or is unrelated to the source text.

    2: The paraphrase has a somewhat different meaning from the source text

    3: The paraphrase captures the general idea of the source text, but some details or nuances are missing.

    4: The paraphrase largely captures the meaning of the source text but may have slight differences in wording or expression.

    5: The paraphrase has an identical or nearly identical meaning to the source text.

    \textbf{Lexical Diversity:} This aspect evaluates the range and richness of vocabulary used in the paraphrase, considering its comparison to the source text.

    \textit{Rating Scale for Lexical Diversity}

    1: The paraphrase shows a limited use of words and lacks diversity when compared to the source text.

    2: The paraphrase exhibits some variation in word choice but heavily relies on a few specific terms, which may not reflect the lexical diversity of the source text.

    3: The paraphrase demonstrates moderate diversity in vocabulary, but there is room for improvement in terms of incorporating more varied word choices from the source text.

    4: The paraphrase displays a good range of vocabulary, utilizing several different words and expressions that align with the lexical diversity of the source text.

    5: The paraphrase showcases an extensive array of vocabulary, demonstrating excellent lexical diversity that closely matches or surpasses the richness of the source text.

    \textbf{Syntactic Diversity:} This aspect assesses the structural variations in the paraphrase compared to the source text.

    \textit{Rating Scale for Syntactic Diversity}

    1: The paraphrase closely mirrors the sentence structure of the source text with minimal variation.

    2: The paraphrase shows some minor changes in sentence structure but largely follows the same pattern as the source text.

    3: The paraphrase introduces moderate variations in sentence structure, deviating from the structure of the source text in certain aspects.

    4: The paraphrase exhibits significant syntactic diversity, using different sentence structures while still conveying the same meaning as the source text.

    5: The paraphrase displays a high level of syntactic diversity, employing various sentence structures creatively while maintaining the meaning of the source text.

    \textbf{Grammatical Correctness:} This evaluates the grammatical accuracy of the paraphrase.

    \textit{Rating Scale for Grammatical Correctness}

    1: The paraphrase contains numerous grammatical errors that significantly impact comprehension.

    2: The paraphrase has several grammatical errors that occasionally affect understanding.

    3: The paraphrase includes some grammatical errors, but they do not hinder overall comprehension.

    4: The paraphrase demonstrates good grammatical correctness with only occasional minor errors.

    5: The paraphrase is grammatically flawless, with no errors or inaccuracies.

    Please provide your ratings for each aspect using the following json format:

    \{"Semantic Similarity": [Rating from 1 to 5],

    "Lexical Diversity": [Rating from 1 to 5],

    "Syntactic Diversity": [Rating from 1 to 5],

    "Grammatical Correctness": [Rating from 1 to 5]\}
    \end{minipage}%
}
\caption{This figure illustrates the prompt fed to the gpt-4 model for evaluation.}
\label{fig:llm-eval-prompt}
\end{figure}

\begin{figure}[h]
    \vspace*{-1cm}
    \makebox[\textwidth][c]{\includegraphics[width=0.57\textwidth]{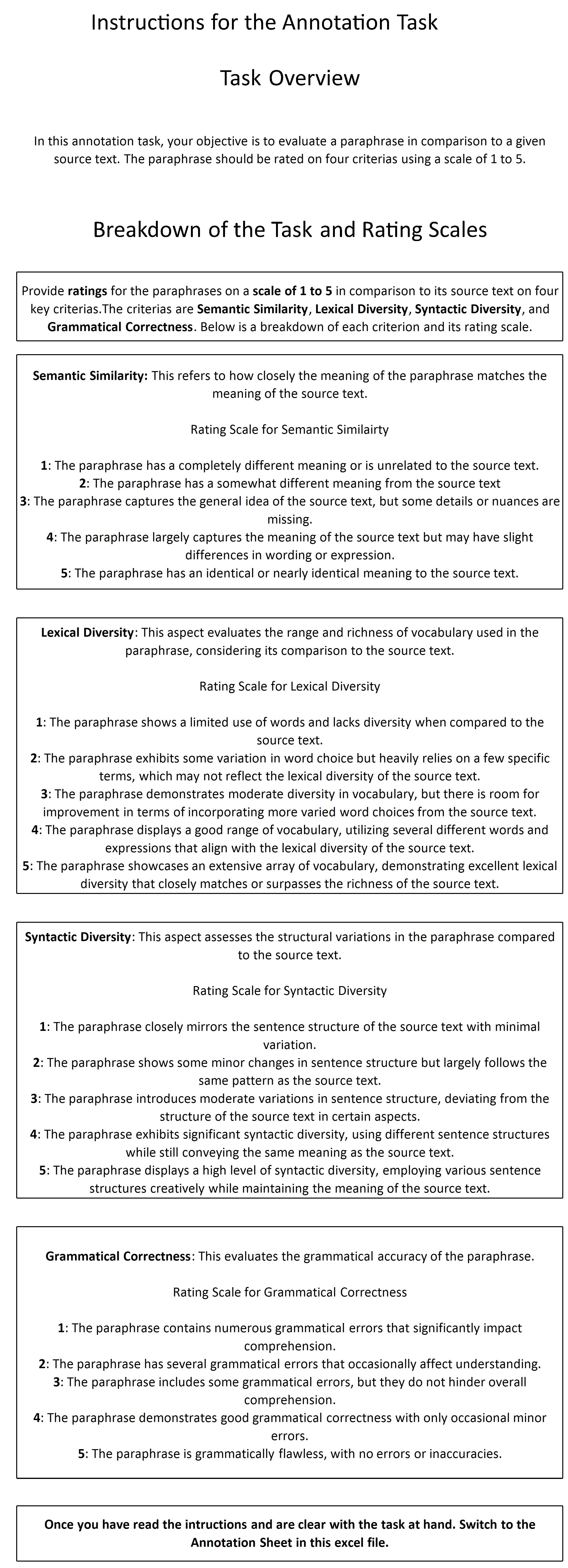}}
    \caption{Instructions given to Human Annotators and breakdown of the 5 point Likert Scale.}
    \label{fig:human-annotator-instructions}
\end{figure}

\end{document}